%% file: paper.tex
\documentclass{article}
                  \usepackage[utf8]{inputenc}
\usepackage[T1]{fontenc}
\usepackage[table,svgnames]{xcolor}
\usepackage[hidelinks]{hyperref}
\usepackage{microtype}
\usepackage{graphicx}
\usepackage{booktabs}
\usepackage{multirow}

\usepackage[accepted]{icml2018}
\usepackage{url}
\usepackage{natbib}
\usepackage{bm}
\usepackage{amsmath}
\usepackage{amssymb}
\usepackage{amsfonts}       
\usepackage{nicefrac}       
\usepackage{mathtools}
\usepackage{tikz}
\usetikzlibrary{chains,colorbrewer}
\usepackage{pgfplots}
\usepgfplotslibrary{colorbrewer}
\usepackage{cleveref}
\usepackage{subcaption}
\renewcommand{\vec}[1]{\bm{#1}}

\newcommand{\vu}{\vec{u}}
\newcommand{\vv}{\vec{v}}
\newcommand{\vx}{\vec{x}}
\newcommand{\vy}{\vec{y}}

\newcommand{\vh}{\vec{h}}

\newcommand{\manifold}[1]{\mathcal{#1}}
\newcommand{\tansp}{\mathcal{T}}

\newcommand{\hyp}{\manifold{L}}
\newcommand{\R}{\mathbb{R}}

\newcommand{\ldot}[1]{\langle #1 \rangle_{\hyp}}
\newcommand{\lnorm}[1]{\| #1 \|_{\hyp}}
\newcommand{\proj}{\text{proj}}
\newcommand{\method}[1]{\textsc{#1}}
\DeclareMathOperator\arcosh{arcosh}
\DeclareMathOperator\grad{grad}
\DeclareMathOperator*{\argmin}{arg\,min}
\newcommand{\sddots}{\raisebox{3pt}{$\scalebox{.75}{$\ddots$}$}}
\newcommand{\cellhi}{\cellcolor{RoyalBlue!10}}
\pgfplotsset{compat=newest}
\date{}
\title{Learning Hierarchical Representations in the Lorentz Model of Hyperbolic Geometry}
\hypersetup{
 pdfauthor={Maximilian Nickel},
 pdftitle={Learning Hierarchical Representations in the Lorentz Model of Hyperbolic Geometry},
 pdfkeywords={},
 pdfsubject={},
 pdfcreator={Emacs 26.1 (Org mode 9.1.13)}, 
 pdflang={English}}
\begin{document}

\twocolumn[
\icmltitlerunning{Learning Continuous Hierarchies in the Lorentz Model of Hyperbolic Geometry}
\icmltitle{Learning Continuous Hierarchies\\ in the Lorentz Model of Hyperbolic Geometry}

\begin{icmlauthorlist}
\icmlauthor{Maximilian Nickel}{fair}
\icmlauthor{Douwe Kiela}{fair}
\end{icmlauthorlist}

\icmlaffiliation{fair}{Facebook AI Research, New York, NY, USA}
\icmlcorrespondingauthor{Maximilian Nickel}{maxn@fb.com}

\vskip 0.3in
]

\printAffiliationsAndNotice{}

\begin{abstract}
We are concerned with the discovery of hierarchical relationships from
large-scale unstructured similarity scores. For this purpose, we study different
models of hyperbolic space and find that learning embeddings in the Lorentz
model is substantially more efficient than in the Poincaré-ball model. We show
that the proposed approach allows us to learn high-quality embeddings of large
taxonomies which yield improvements over Poincaré embeddings,
especially in low dimensions. Lastly, we apply our model to discover hierarchies
in two real-world datasets: we show that an embedding in hyperbolic space can
reveal important aspects of a company's organizational structure as well as reveal
historical relationships between language families.
\end{abstract}

\section{Introduction}
\label{sec:org601e89b}
Hierarchical structures are ubiquitous in knowledge representation and
reasoning. For example, starting with Linnaeus, taxonomies have long been used
in biology to categorize and understand the relationships between species
\cite{mayr1968role}. In social science, hierarchies are used to understand
interactions in humans and animals or to analyze organizational structures such
as companies and governments \citep{networks/dodds2003information}. In comparative
linguists, evolutionary trees are used to describe the origin of languages
\citep{campbell2013historical}, while ontologies are used to provide rich
categorizations of entities in semantic networks \citep{antoniou2004web}.
Hierarchies are also known to provide important information for learning and
classification \citep{silla2011survey}. In cognitive development, the results of
\citet{inhelder1964growth} suggest that the classification structure in children's
thinking is hierarchical in nature.

Hierarchies can therefore provide important insights into systems of concepts.
However, explicit information about such hierarchical relationships is
unavailable for many domains. In this paper, we therefore consider the problem
of discovering concept hierarchies from unstructured observations, specifically
in the following setting:

\begin{enumerate}
\item We focus on discovering pairwise hierarchical relations between concepts,
where all superior and subordinate concepts are observed.

\item We aim to infer concept hierarchies only from pairwise similarity
measurements, which are relatively easy and cheap to obtain in many domains.
\end{enumerate}

Examples of hierarchy discovery that adhere to this setting include the creation
of taxonomies from similarity judgments (e.g., genetic similarity of species or
cognate similarity of languages) and the recovery of organizational hierarchies
and dominance relations from social interactions.

To infer hierarchies from similarity judgments, we propose to model such
relationships as a combination of two separate aspects: relatedness and
generality. Concept A is a parent (a superior) to concept B if both concepts are
related and A is more general than B. By separating these aspects, we can then
discover concept hierarchies via hyperbolic embeddings. In particular, we build
upon ideas of Poincaré embeddings \citep{nickel2017poincare} to learn continuous
representations of hierarchies. Due to its geometric properties, hyperbolic
space can be thought of as continuous analogue to discrete trees. By embeddings
concepts in such a way that their similarity order is preserved, we can then
identify (soft) hierarchical relationships from the embedding: relatedness is
captured via the distance in the embedding space, while generality is captured
via the norm of the embeddings.

To learn high-quality embeddings, we propose a new optimization approach based
on the Lorentz model of hyperbolic space. The Lorentz model allows for an
efficient closed-form computation of the geodesics on the manifold. This
facilitates the development of an efficient optimizer that directly follows
these geodesics, rather than doing a first-order approximation as in
\citep{nickel2017poincare}. It allows us also to avoid numerical instabilities
that arise from the Poincaré distance. As we will show experimentally, this
optimization method leads to a substantially improved embedding quality,
especially in low dimensions. Simultaneously, we retain the attractive
properties of hyperbolic embeddings, i.e., learning continuous representations
of hierarchies via gradient-based optimization while scaling to large datasets.

The reminder of this paper is organized as follows. In \Cref{sec:related}, we
discuss related work regarding hyperbolic and ordered embeddings. In
\Cref{sec:related}, we introduce our model and algorithm to compute the
embeddings. In \Cref{sec:experiments} we evaluate the efficiency of our approach on
large taxonomies. Furthermore, we evaluate the ability of our model to discover
meaningful hierarchies on real-world datasets.

\section{Related Work}
\label{sec:related}
Hyperbolic geometry has recently received attention in machine learning and
network science due to its attractive properties for modeling data with latent
hierarchies. \citet{hyperbolic/krioukov2010hyperbolic} showed that typical
properties of complex networks (e.g., heterogeneous degree distributions and
strong clustering) can be explained by assuming an underlying hyperbolic
geometry and, moreover, developed a framework to model networks based on these
properties. Furthermore, \citet{kleinberg2007geographic} and
\citet{boguna2010sustaining} proposed hyperbolic embeddings for greedy
shortest-path routing in communication networks. \citet{asta2015geometric} used
hyperbolic embeddings of graphs to compare the global structure of networks.
\citet{sun2015spacetime} proposed to learn representations of non-metric data in
pseudo-Riemannian space-time, which is closely related to hyperbolic space.

Most similar to our work are the recently proposed Poincaré embeddings
\citep{nickel2017poincare}, which learn hierarchical representations of symbolic
data by embedding them into an \(n\)-dimensional Poincaré ball. The main focus
of that work was to model the link structure of symbolic data efficiently, i.e.,
to find low-dimensional embeddings via exploiting the hierarchical structure of
hyperbolic space. Here, we build upon this idea and extend it in various ways.
First, we propose a new model to compute hyperbolic embeddings in the Lorentz
model of hyperbolic geometry. This allows us to develop an efficient Riemannian
optimization method that scales well to large datasets and provides
better embeddings, especially in low dimensions. Second, we
consider inferring hierarchies from real-valued similarity scores, which
generalize binary adjacency matrices as considered by \citet{nickel2017poincare}.
Third, in addition to preserving similarity (e.g., local link structure), we
also focus on recovering the correct hierarchical relationships from the embedding.

Simultaneously to the present work, \citet{de2018representation} analyzed the
representation trade-offs for hyperbolic embeddings and proposed a new
combinatorial embedding approach as well as a new approach to Multi-Dimensional
Scaling (MDS) in hyperbolic space. Furthermore, \citet{ganea2018hyperbolic}
extended Poincaré embeddings using geodesically convex cones to model
asymmetric relations.

Another related method is Order Embeddings \citep{vendrov2015order}, which was
proposed to learn visual-semantic hierarchies over words, sentences, and images
from ordered input pairs. In contrast, we are concerned with learning
hierarchical embeddings from less supervision: namely, from \emph{unordered}
(symmetric) input pairs that provide no direct information about the partial
ordering in the hierarchy. 

Further work on embedding order-structures include
Stochastic Triplet Embeddings \citep{van2012stochastic}, Generalized Non-Metric
MDS \citep{agarwal2007generalized}, and Crowd Kernels \citep{tamuz2011adaptively}.
In the context of word embeddings, \citet{vilnis2015word} proposed Gaussian
Embeddings to learn improved representations. By mapping words to densities,
this model is capable of capturing uncertainty, assymmetry, and (hierarchical)
entailment relations.

To discover structural forms (e.g., trees, grids, chains) from data,
\citet{kemp2008discovery} proposed a model for making probabilistic inferences
over a space of graph grammars. Recently, \citet{lake2018emergence} proposed an
alternative approach to this work based on structural sparsity. Additionally,
hierarchical clustering has a long history in machine learning and data mining
\citep{duda1973pattern}. Bottom-up agglomerative clustering assigns each data
point to its own cluster and then iteratively merges the two closest points
according to a given distance measure (e.g., single link, average link, max
link). As such, hierarchical clustering provides a hierarchical partition of the
input space. In contrast, we are concerned with discovering direct hierarchical
relationships between the input data points.

\section{Methods}
\label{sec:methods}
In the following, we describe our approach for learning continuous hierarchies
from unstructured observations.

\begin{figure*}[t]
  \centering
  \begin{minipage}{0.45\linewidth}
    \centering
    \vspace{1em}
    \resizebox{0.7\linewidth}{!}{\input{geodesics.pgf}}
    \vspace{1em}
    \subcaption{Geodesics in the Poincaré disk.}\label{fig:geodesics}
  \end{minipage}
  \hfill
  \begin{minipage}{0.45\linewidth}
    \centering
    \includegraphics[trim={2cm 7cm 2cm
      7cm},clip,width=\columnwidth]{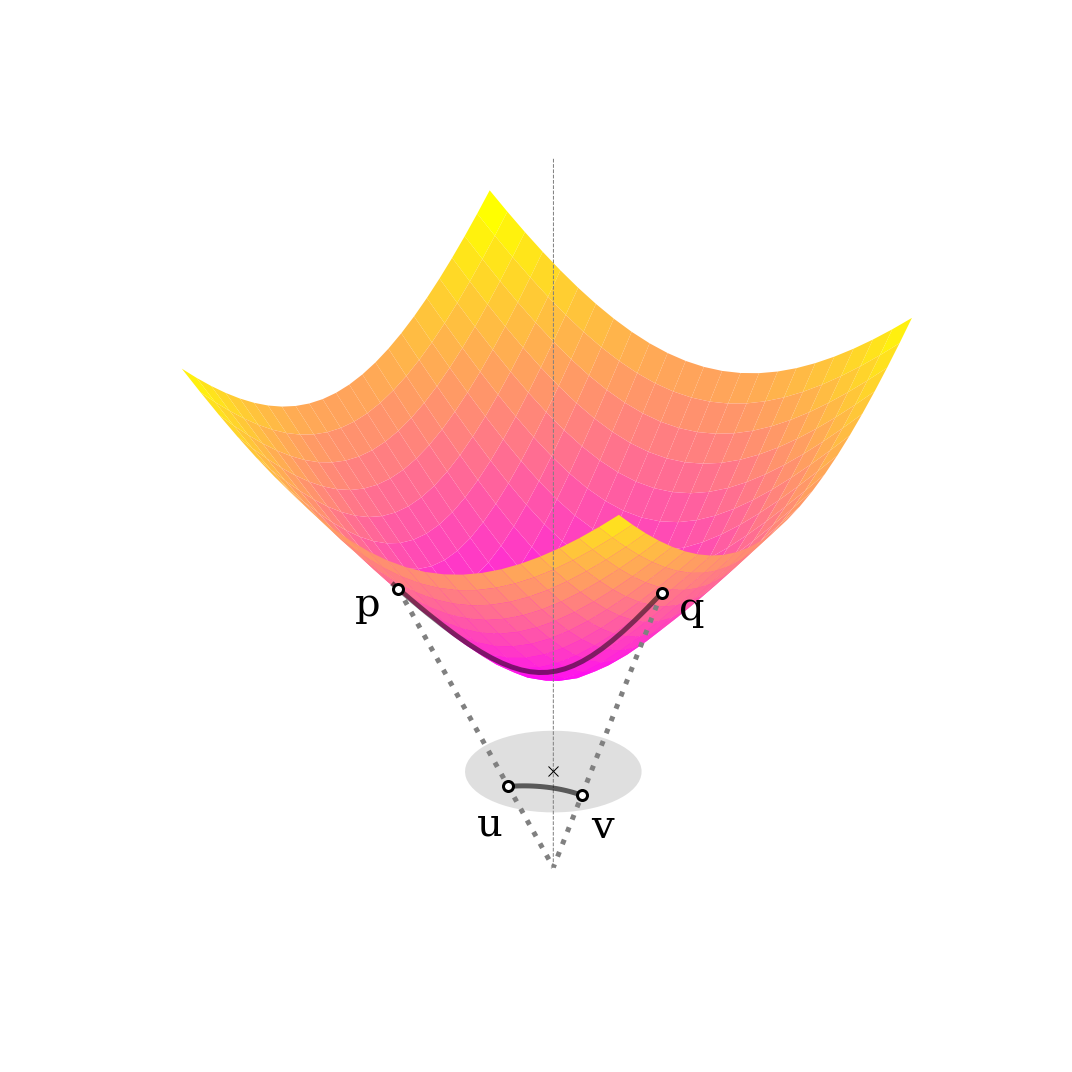}
    \subcaption{Lorentz model of hyperbolic geometry.}\label{fig:hyperboloid}
  \end{minipage}
  \caption{\subref{fig:geodesics}) Geodesics in the Poincaré disk model of hyperbolic space. Due to the negative curvature of the space, geodesics between points are arcs that are perpendicular to the boundary of the disk.
    For curved arcs, midpoints are closer to the origin of the disk (p1) than
    the associated points, e.g. (p3, p5). \subref{fig:hyperboloid}) Points (p,q)
    lie on the surface of the upper sheet of a two-sheeted hyperboloid.
    Points (u, v) are the mapping of (p, q) onto the Poincaré disk using \Cref{eq:p-to-h}.}
\end{figure*}

\subsection{Hyperbolic Geometry \& Poincaré Embeddings}
\label{sec:orgcde4f33}
Hyperbolic space is the unique, complete, simply connected Riemannian manifold
with constant negative sectional curvature. There exist multiple
equivalent\footnote{Meaning that there exist transformations between the different
models that preserve all geometric properties including isometry.} models for
hyperbolic space and one can choose the model whichever is best suited for a
given task. \citet{nickel2017poincare} based their approach for learning
hyperbolic embeddings on the Poincaré ball model, due to its conformality and
convenient parameterization. The Poincaré ball model is the Riemannian manifold
\(\manifold{P}^n = (\manifold{B}^n, g_p)\), where \(\manifold{B}^n = \{\vx \in \R^n : \|\vx\| < 1\}\)
is the \emph{open} \(n\)-dimensional unit ball and where
\begin{equation*}
g_p(\vx) = \left(\frac{2}{1 - \|\vx\|^2}\right)^2 g_e
\end{equation*}
The distance function on \(\manifold{P}\) is then defined as
\begin{equation}
d_p(\vx, \vy) = \arcosh \left(1 + 2 \frac{\|\vx - \vy\|^2}{(1 - \|\vx\|^2)(1 - \|\vy\|^2)} \right)
\label{eq:pdist}
\end{equation}

It can be seen from \Cref{eq:pdist}, that the distance within the Poincaré ball
changes smoothly with respect to the norm of \(\vx\) and \(\vy\). This locality
property of the distance is key for learning continuous embeddings of
hierarchies. For instance, by placing the root node of a tree at the origin of
\(\manifold{B}^n\), it would have relatively small distance to all other nodes, as
its norm is zero. On the other hand, leaf nodes can be placed close to the
boundary of the ball, as the distance between points grows quickly with a norm
close to one.

\subsection{Riemannian Optimization in the Lorentz Model}
\label{sec:roptim}
In the following, we propose a new method to compute hyperbolic embeddings based
on the Lorentz model of hyperbolic geometry. The main advantage of this
parameterization is that it allows us to perform Riemannian optimization very
efficiently. An additional advantage is that its distance function (see
\Cref{eq:ldist}) avoids numerical instabilities that arise from the fraction in the
Poincaré distance.

\subsubsection{The Lorentz Model of Hyperbolic space}
\label{sec:org8479d36}
In the following, let \(\vx\), \(\vy \in \R^{n+1}\) and let
\begin{equation}
  \ldot{\vx, \vy} = -x_0 y_0 + \sum_{i=1}^n x_n y_n
\end{equation}
denote the \emph{Lorentzian scalar product}. The Lorentz model of
\(n\)-dimensional hyperbolic space is then defined as the Riemannian
manifold \(\mathcal{L}^n = (\manifold{H}^n, g_\ell)\), where 
\begin{equation}
  \manifold{H}^n = \{\vx \in \R^{n+1} : \ldot{\vx, \vx} = -1, x_0 > 0\}
\end{equation}
denotes the upper sheet of a two-sheeted \(n\)-dimensional hyperboloid and where
\begin{equation}
  g_\ell(\vx) = \begin{bsmallmatrix} -1 \\ & 1 \\ & & \sddots \\ & & & 1 \end{bsmallmatrix} .
\end{equation}

The associated distance function on \(\mathcal{L}\) is then given as
\begin{equation}
  d_\ell(\vx, \vy) = \arcosh(-\ldot{\vx, \vy}) \label{eq:ldist}
\end{equation}
Furthermore, it holds for any point \(\vx = (x_0, \vx^\prime) \in \R^{n+1}\)
\begin{equation}
  \vx \in \manifold{H}^n \Leftrightarrow x_0 = \sqrt{1 + \|\vx^\prime\|} \label{eq:hvec}
\end{equation}

\subsubsection{Riemannian Optimization}
\label{sec:orgcb47a00}
To derive the Riemannian SGD (RSGD) algorithm for the Lorentz model, we will
first review the necessary concepts of Riemannian optimization. A Riemannian
manifold \((\manifold{M}, g)\) is a real, smooth manifold \(\manifold{M}\)
equipped with a Riemannian metric \(g\). Furthermore, for each \(\vx \in
\manifold{M}\), let \(\tansp_{\vx}\manifold{M}\) denote the associated \emph{tangent
space}. The metric \(g\) induces then a inner product \(\langle \cdot, \cdot
\rangle_{\vx} : \tansp_{\vx}\manifold{M} \times \tansp_{\vx}\manifold{M} \to
\R\). \emph{Geodesics} \(\gamma : [0, 1] \to \manifold{M}\) are the generalizations
of straight lines to Riemannian manifolds, i.e., constant speed curves that
are locally distance minimizing. The \emph{exponential map} \(\exp_{\vx} :
\tansp_{\vx}\manifold{M} \to \manifold{M}\) maps a tangent vector \(\vv \in
\tansp_{\vx}\manifold{M}\) onto \(\manifold{M}\) such that \(\exp_{\vx}(\vv) =
\vy\), \(\gamma(0) = \vx\), \(\gamma(1) = \vy\) and \(\dot{\gamma}(0) =
\frac{\partial}{\partial t} \gamma(0) = \vv\). For a \emph{complete manifold}
\(\manifold{M}\), the exponential map is defined for all points \(\vx \in
\manifold{M}\).

Furthermore, let \(f : \manifold{M} \to \R\) be a smooth real-valued function over
parameters \(\theta \in \manifold{M}\). In Riemannian optimization, we are
then interested in solving problems of the form
\begin{equation}
  \min_{\theta \in \manifold{M}} f(\theta) .\label{eq:roptim}
\end{equation}

Following \citet{optimization/bonnabel2013stochastic}, we minimize
\Cref{eq:roptim} using Riemannian SGD. In RSGD, updates to the parameters
\(\theta\) are computed via
\begin{equation}
  \theta_{t+1} = \exp_{\theta_t}(-\eta \grad f(\theta_t))
\end{equation}
where \(\grad f (\theta_t) \in \tansp_\theta \manifold{M}\) denotes the
\emph{Riemannian gradient} and \(\eta\) denotes the learning rate.

For the Lorentz model, the tangent space is defined as follows: For a point \(\vx
\in \hyp^n\), the tangent space \(\tansp_{\vx}\hyp^n\) consists of
all vectors orthogonal to \(\vx\), where orthogonality is defined with respect to
the Lorentzian scalar product. Hence,
\begin{equation*}
  \tansp_{\vx} \hyp^n = \{\vv : \ldot{\vx, \vv} = 0 \} .
\end{equation*}

Furthermore, let \(\vv \in \tansp_{\vx} \hyp^n\). The exponential map
\({\exp_{\vx} : \tansp_{\vx} \hyp^n \to \hyp^n}\) is then
defined as
\begin{equation}
  \exp_{\vx}(\vv) = \cosh(\lnorm{\vv})\vx + \sinh(\lnorm{\vv})\frac{\vv}{\lnorm{\vv}}
  \label{eq:expm}
\end{equation}
where \(\lnorm{\vv} = \sqrt{\ldot{\vv,\vv}}\) denotes the norm of \(\vv\) in \(\tansp_{\vx} \hyp^n\).

To compute parameter updates as in \Cref{eq:roptim}, we need additionally the
Riemannian gradient of \(f\) at \(\theta\). For this purpose, we first compute
the direction of steepest descent from the Euclidean gradient \(\nabla
f(\theta)\) via
\begin{equation}
  \vec{h} = g_\ell^{-1}\nabla f(\theta)
  \label{eq:direction}
\end{equation}
Since \(g_\ell\) is an involutory matrix (i.e., \(g_\ell^{-1} = g_\ell\)), the
inverse in \Cref{eq:direction} is trivial to compute. To derive the Riemannian
gradient from \(\vec{h} \in \R^{n+1}\), we then use the orthogonal projection
\(\proj_\theta : \R^{n+1} \to \tansp_\theta \hyp^n\) from the ambient
Euclidean space onto the tangent space of the current
parameter. This projection is computed as 
\begin{equation*}
  \proj_{\vx}(\vu) = \vu - \frac{\ldot{\vx, \vu}}{\ldot{\vx, \vx}} \vx = \vu + \ldot{\vx, \vu} \vx
  \label{eq:proj}
\end{equation*}
since \(\forall x \in \manifold{H}^n : \ldot{\vx, \vx} = -1\)
\cite{hyperbolic/robbin2017introduction}. Using \Cref{eq:expm,eq:proj}, we can
then estimate the parameters \(\theta\) using RSGD as in \Cref{alg:rsgd}. We initialize
the embeddings close to the origin of \(\manifold{H}^n\) by sampling from the uniform distribution \(\mathcal{U}(-0.001, 0.001)\) and by setting \(x_0\)
according to \Cref{eq:hvec}.

\begin{algorithm}[t]
  \caption{Riemannian Stochastic Gradient Descent \label{alg:rsgd}}
  \begin{tabular}{ll}
    \multicolumn{2}{l}{\textbf{Input} Learning rate $\eta$, number of epochs $T$.}\\
    \multicolumn{2}{l}{for $t=1,\ldots,T$}\\
    $\quad \vh_t$ & $\gets g_{\theta_t}^{-1} \nabla f(\theta_t)$ \\
    $\quad \grad f(\theta_t)$ & $ \gets \proj_{\theta_{t}}(\vh_t)$ \\
    $\quad \theta_{t+1}$ & $\gets \exp_{\theta_{t}}(-\eta \grad f(\theta_t))$
  \end{tabular}
\end{algorithm}

\subsubsection{Equivalence of models}
\label{sec:org04ad244}
The Lorentz and Poincaré disk model both have specific strengths: the Poincare
disk provides a very intuitive method for visualizing and interpreting
hyperbolic embeddings. The Lorentz model on the other hand is well-suited for
Riemannian optimization. Due to the equivalence of both models, we can exploit
their individual strengths simultaneously: points in the Lorentz
model can be mapped into the Poincaré ball via the diffeomorphism \(p :
\manifold{H}^n \to \manifold{P}^n\), where
\begin{equation}
  p(x_0, x_1, \ldots, x_n) = \frac{(x_1, \ldots, x_n)}{x_0 + 1}
  \label{eq:p-to-h}
\end{equation}

Furthermore, points in \(\manifold{P}^n\) can be mapped into \(\manifold{H}^n\) via
\begin{equation*}
  p^{-1}(x_1, \ldots, x_n) = \frac{(1 + \|x\|^2, 2x_1,\ldots,2x_n)}{1 - \|x\|^2}
\end{equation*}

We will therefore learn the embeddings via \Cref{alg:rsgd} in the Lorentz model
and visualize the embeddings by mapping them into the Poincaré disk using
\Cref{eq:p-to-h}. See also \Cref{fig:hyperboloid} for an illustration of Lorentz model
and its connections to the Poincaré disk.

\subsection{Inferring Concept Hierarchies from Similarity}
\label{sec:orgec11ac9}
\citet{nickel2017poincare} embedded unweighted undirected
graphs in hyperbolic space. In the following, we extend this approach to a more general
setting, i.e., inferring continuous hierarchies from \emph{pairwise
similarity measurements}.

Let \(\mathcal{C} = \{c_i\}_{i=1}^m\) be a set of concepts and \(X \in \R^{m \times
m}\) be a dataset of pairwise similarity scores between these concepts. We also
assume that the concepts can be organized according to an unobserved hierarchy
\((\mathcal{C}, \preceq)\), where \({c_i \preceq c_j}\) defines a \emph{partial order}
over the elements of \(\mathcal{C}\). Since partial order is a reflexive,
anti-symmetric, and transitive binary relation, it is well suited to define
hierarchical relations over \(\mathcal{C}\). If \(c_i \preceq c_j\) or
\(c_j \preceq c_i\), then the concepts \(c_i\), \(c_j\) are
\emph{comparable} (e.g., located in the same subtree). Otherwise they are
incomparable (e.g., located in different subtrees). For concepts \(c_i \prec
c_j\), we will refer to \(c_i\) as the superior and to \(c_j\) as the subordinate
node.

Given this setting, our goal is then to recover the partial order \((\mathcal{C},
\preceq)\) from \(X\). For this purpose, we separate the semantics of the partial
order relation into two distinct aspects: First, whether two concepts \(c_i, c_j\)
are comparable (denoted by \(c_i \sim c_j\)) and, second, whether concept \(c_i\) is
more \emph{general} than \(c_j\) (denoted by \(c_j \sqsubset c_i\)). Combining both
aspects provides us with the usual interpretation of partial order.

By explicitly distinguishing between the aspects of comparability and
generality, we can then make the following structural assumptions on \(X\) to
infer hierarchies from pairwise similarities: 1) Comparable (and related)
concepts are more similar to each other than incomparable concepts (i.e., \(X_{ij}
\geq X_{ik}\) if \(c_i \preceq c_j \land c_i \not \preceq c_k\)); and 2) We assume that
general concepts are similar to more concepts than less general ones. Both are
mild assumptions given that the similarity scores \(X\) describe concepts that are
organized in a latent hierarchy. For instance, 1) simply follows from the
assumption that concepts in the same subtree of the ground-truth hierarchy
are more similar to each other than to concepts in different subtrees. This is also
used in methods that use path-lengths in taxonomies to measure the \emph{semantic
similarity} of concepts (e.g., see \citealp{resnik1999semantic}).

It follows from assumption 1) the we want to preserve the similarity orderings
in the embedding space in order to predict comparability. In particular, let
\(\vu_i\) denote the embedding of \(c_i\) and let \(\mathcal{N}(i,j) = \{k : X_{ik} <
X_{ij}\} \cup \{j\}\) denote the set of concepts that are \emph{less} similar to \(c_i\)
then \(c_j\) (including \(c_j\)). Based only on pairwise similarities in \(X\), it is
difficult to make global decisions about the likelihood that \(c_i \sim c_j\) is
true. However, it follows from assumption 1) that we can make \emph{local ranking}
decisions, i.e., we can infer that \(c_i \sim c_j\) is the most likely among all
\(c_k \in \mathcal{N}(i, j)\). For this purpose, let
\begin{equation*}
  \phi(i, j) = \argmin_{k \in \mathcal{N}(i, j)} d(\vu_i, \vu_k)
\end{equation*}
be the nearest neighbor of \(c_i\) in the set \(\mathcal{N}(i,j)\). We then learn
embeddings \(\Theta = \{\vu\}_{i=1}^m\) by optimizing
\begin{equation}
  \max_{\Theta} \sum_{i,j} \log \Pr(\phi(i,j) = j\ |\ \Theta)
  \label{eq:loss}
\end{equation}
where
\begin{equation*}
  \Pr(\phi(i,j) = j\ |\ \Theta)= \frac{e^{-d(\vu_i, \vu_j)}}{\sum_{k \in \mathcal{N}(i,j)} e^{-d(\vu_i, \vu_k)}}
\end{equation*}
For computational efficiency, we follow \cite{jean2014using} and randomly
subsample \(\mathcal{N}(i,j)\) on large datasets.

\Cref{eq:loss} is a ranking loss that aims to preserve the neighborhood structures
in \(X\). For each pair of concepts \((i,j)\), this loss induces embeddings where
\((i,j)\) is closer in the embedding space than pairs \((i,k)\) that are less
similar. Since we compute the embedding in a metric space, we also retain
transitive relations approximately. We can therefore identify the comparability
of concepts \(c_i \sim c_j\) by their distance \(d(\vu_i, \vu_j)\) in the embedding.

Moreover, by optimizing \Cref{eq:loss} in hyperbolic space, we are also able to
infer the generality of concepts from their embeddings. According to assumption
2), we can can assume that general objects will be close to many different
concepts. Since \Cref{eq:loss} optimizes the local similarity ranking for all
concepts, we can also assume that this ordering is preserved. We can see from
\Cref{eq:pdist} that points with a small distance to many different points are
located close to the center. We can therefore identify the generality of a
concept \(c_i\) simply via the norm of its embedding \(\|\vu_i\|\).

We have now cast the problem of hierarchy discovery as a simple embedding
problem whose objective is to preserve local similarity orderings

\section{Evaluation}
\label{sec:experiments}
\begin{table}[t]
\caption{\label{tab:tax-stats}
Taxonomy Statistics. The number of edges refers to the full transitive closure of the respective taxonomy.}
\centering
\small
\begin{tabular}{lrrr}
\toprule
\textbf{Taxonomy} & \textbf{Nodes} & \textbf{Edges} & \textbf{Depth}\\
\midrule
\method{WordNet} Nouns & 82,115 & 769,130 & 19\\
\method{WordNet} Verbs & 13,542 & 35,079 & 12\\
\method{EuroVoc} (en) & 7,084 & 10,547 & 5\\
ACM & 2,299 & 6,526 & 5\\
\method{MeSH} & 28,470 & 191,849 & 15\\
\bottomrule
\end{tabular}
\end{table}

\begingroup
\setlength{\tabcolsep}{5pt}
\begin{table*}
  \centering
  \small
  \caption{Evaluation of Taxonomy Embeddings. MR = Mean Rank,
    MAP = Mean Average Precision $\rho$ = Spearman rank-order correlation. $\Delta\%$ indicates the
    relative improvement of optimization in the Lorentz model.\label{tab:tax}}
  \vspace{1ex}
  \begin{tabular}{llccccccccccccccc}
    \toprule
    & & 
        \multicolumn{3}{c}{\bf \method{WordNet} Nouns} & 
                                                         \multicolumn{3}{c}{\bf \method{WordNet} Verbs} & 
                                                                                                          \multicolumn{3}{c}{\bf \method{EuroVoc}} & 
                                                                                                                                                     \multicolumn{3}{c}{\bf \method{ACM}} & 
                                                                                                                                                                                            \multicolumn{3}{c}{\bf \method{MeSH}}\\
    \cmidrule(r){3-5}
    \cmidrule(lr){6-8}
    \cmidrule(lr){9-11}
    \cmidrule(r){12-14}
    \cmidrule(r){15-17}
    & & 2 & 5 & 10 & 2 & 5 & 10 & 2 & 5 & 10 & 2 & 5 & 10 & 2 & 5 & 10\\
    \midrule
    \multirow{3}{*}{\bf MR} 
    & Poincaré & \cellhi 90.7 & 4.9 & 4.02 & \cellhi 10.71 & 1.39 & 1.35 & \cellhi 2.83 & 1.25 & 1.23 & \cellhi 4.14 & 1.8 & 1.71 & \cellhi 61.11 & 14.05 & 12.8 \\
    & Lorentz & \cellhi 22.8 & 3.18 & 2.95 & \cellhi 3.64 & 1.26 & 1.23 & \cellhi 1.63 & 1.24 & 1.17 & \cellhi 3.05 & 1.67 & 1.63 & \cellhi 38.99 & 14.13 & 12.42 \\
    & $\Delta\%$ & \cellhi 74.8 & 35.1 & 36.2 & \cellhi 66.0 & 9.6 & 8.9 & \cellhi 42.4 & 6.1 & 3.4 & \cellhi 26.3 & 7.2 & 4.8 & \cellhi 36.2 & -0.5 & 2.9 \\
    \midrule
    \multirow{3}{*}{\bf MAP} 
    & Poincaré & \cellhi 11.8 & 82.8 & 86.5 & \cellhi 36.5 & 91.0 & 91.2 & \cellhi 64.3 & 94.0 & 94.4 & \cellhi 69.3 & 94.1 & 94.8 & \cellhi 19.5 & 76.3 & 79.4\\
    & Lorentz & \cellhi 30.5 & 92.3 & 92.8 & \cellhi 57.9 & 93.5 & 93.3 & \cellhi 87.1 & 95.8 & 96.5 & \cellhi 82.9 & 96.6 & 97.0 & \cellhi 34.8 & 77.7 & 79.9 \\
    & $\Delta\%$ & \cellhi 61.3 & 10.3 & 6.8 & \cellhi 58.6 & 2.7 & 2.3 & \cellhi 35.6 & 1.6 & 2.0 & \cellhi 19.6 & 2.7 & 2.3 & \cellhi 43.9 & 1.8 & 0.6\\
    \midrule
    \multirow{2}{*}{$\bm{\rho}$} 
    & Poincaré & \cellhi 13.8 & 57.2 & 58.5 & \cellhi 11.0 & 54.1 & 55.1 & \cellhi 37.5 & 57.5 & 61.4 & \cellhi 59.8 & 63.5 & 62.9 & \cellhi 42.2 & 69.9 & 74.9\\
    & Lorentz & \cellhi 41.0 & 58.9 & 59.5 & \cellhi 47.9 & 55.5 & 56.6 & \cellhi 54.5 & 61.7 & 67.5 & \cellhi 65.9 & 65.9 & 65.9 & \cellhi 64.5 & 71.4 & 76.3 \\
    \bottomrule
  \end{tabular}
\end{table*}
\endgroup

\subsection{Embedding Taxonomies}
\label{sec:exp:tax}
In the following experiments, we evaluate the performance of the Lorentz model for embedding large taxonomies. For this purpose, we compore its embedding quality to Poincaré embeddings \citep{nickel2017poincare} on the following real-world taxonomies

\begin{description}
\item[{WordNet \textregistered}] \citep{miller1998wordnet} is a large lexical database which, amongst other relations, provides hypernymy (is-a) relations. In our experiments, we embedded the noun and verb hierarchy of WordNet.
\item[{EuroVoc}] is a mulitlingual thesaurus maintained by the European Union. It contains keywords organized in 21 domains and 127 sub-domains. In our experiments, we used the English section of EuroVoc.\footnote{Available at \url{http://eurovoc.europa.eu}}
\item[{ACM}] The ACM computing classification system is a hierarchical ontology which is used by various ACM journals to organize subjects by area.
\item[{MeSH}] Medical Subject Headings (MeSH; \cite{rogers1963medical}) is a medical thesaurus which is created, maintained and provided by the U.S. National Library of Medicine. In our experiments we used the 2018 MeSH hierarchy.
\end{description}
Statistics for all taxonomies are provided in \Cref{tab:tax-stats}. 

In our evaluation, we closely follow the setting of \citet{nickel2017poincare}: First,
we embed the undirected transitive closure of these taxonomies, such that the
hierarchical structure is not directly visible from the observed edges but has
to be inferred. To measure the quality of the embedding, we compute
for each observed edge \((u, v)\) the corresponding distance \(d(\vu, \vv)\) in the
embedding and rank it among the distances of all \emph{unobserved} edges for \(u\),
i.e., among \(\{d(\vu, \vv^\prime) : (u, v^\prime) \not \in \mathcal{D}\}\). We
then report the mean rank (MR) and mean average precision (MAP) of this
ranking.

In addition, we also evaluate how well the norm of the embeddings (i.e., our
indicator for generality), correlates with the ground-truth ranks in the
embedded taxonomy. Since different subtrees can have very different depths, we
normalize the rank of each concept by the depth of its subtree and measure the
Spearman rank-order correlation \(\rho\) of the normalized rank with the norm of
the embedding. We compute the normalized rank in the following way: Let
\(\text{sp}(c)\) denote the shortest path to the root node from \(c\), and let
\(\text{lp}(c)\) denote the longest path from \(c\) to any of its
children.\footnote{Since all taxonomies in our experiments are DAGs, it is possible
to compute the longest path in the graph} The normalized rank of \(c\) is the given as
\begin{equation*}
  \text{rank}(c) = \frac{\text{sp}(c)}{\text{sp}(c) + \text{lp}(c)} .
\end{equation*}

To learn the embeddings in the Lorentz model, we employ the Riemannian
optimization method as described in \Cref{sec:roptim}. For Poincaré embeddings, we
use the official open-source
implementation.\footnote{Source code available at \url{https://github.com/facebookresearch/poincare-embeddings}}
Both methods were cross-validated over identical sets of hyperparameters.

\Cref{tab:tax} shows the results of our evaluation. It can be seen that both
methods are very efficient in embedding these large taxonomies. However, the
Lorentz model shows consistently higher-quality embeddings and especially so in
low dimensions. The relative improvement of the two-dimensional Lorentz
embeddings over the Poincaré embedding amounts to 74.8\% on the \method{WordNet}
noun hierarchy and 42.4\% on \method{EuroVoc}. Similar improvements can be
observed on all taxonomies. Furthermore, on the most complex taxonomy
(\method{WordNet} nouns), the 10-dimensional Lorentz embeddings already
out-performs the best reported numbers reported in \citep{nickel2017poincare}
(which went up to 200 dimensions). This suggests that the full Riemannian
optimization approach can be very helpful for obtaining good embeddings. This is
especially the case in low dimensions where it is harder for the optimization
procedure to escape from local minima.

\subsection{Enron Email Corpus}
\label{sec:orgbb8b8bf}
\definecolor{Accent-6-1}{RGB}{127,201,127}
\definecolor{Accent-6-A}{RGB}{127,201,127}
\definecolor{Accent-6-2}{RGB}{190,174,212}
\definecolor{Accent-6-B}{RGB}{190,174,212}
\definecolor{Accent-6-3}{RGB}{253,192,134}
\definecolor{Accent-6-C}{RGB}{253,192,134}
\definecolor{Accent-6-4}{RGB}{255,255,153}
\definecolor{Accent-6-D}{RGB}{255,255,153}
\definecolor{Accent-6-5}{RGB}{56,108,176}
\definecolor{Accent-6-E}{RGB}{56,108,176}
\definecolor{Accent-6-6}{RGB}{240,2,127}
\definecolor{Accent-6-F}{RGB}{240,2,127}

\definecolor{YlGnBu-6-1}{RGB}{255,255,204}
\definecolor{YlGnBu-6-B}{RGB}{255,255,204}
\definecolor{YlGnBu-6-2}{RGB}{199,233,180}
\definecolor{YlGnBu-6-D}{RGB}{199,233,180}
\definecolor{YlGnBu-6-3}{RGB}{127,205,187}
\definecolor{YlGnBu-6-F}{RGB}{127,205,187}
\definecolor{YlGnBu-6-4}{RGB}{65,182,196}
\definecolor{YlGnBu-6-G}{RGB}{65,182,196}
\definecolor{YlGnBu-6-5}{RGB}{44,127,184}
\definecolor{YlGnBu-6-I}{RGB}{44,127,184}
\definecolor{YlGnBu-6-6}{RGB}{37,52,148}
\definecolor{YlGnBu-6-K}{RGB}{37,52,148}

\definecolor{YlGnBu-9-1}{RGB}{255,255,217}
\definecolor{YlGnBu-9-A}{RGB}{255,255,217}
\definecolor{YlGnBu-9-2}{RGB}{237,248,177}
\definecolor{YlGnBu-9-C}{RGB}{237,248,177}
\definecolor{YlGnBu-9-3}{RGB}{199,233,180}
\definecolor{YlGnBu-9-D}{RGB}{199,233,180}
\definecolor{YlGnBu-9-4}{RGB}{127,205,187}
\definecolor{YlGnBu-9-F}{RGB}{127,205,187}
\definecolor{YlGnBu-9-5}{RGB}{65,182,196}
\definecolor{YlGnBu-9-G}{RGB}{65,182,196}
\definecolor{YlGnBu-9-6}{RGB}{29,145,192}
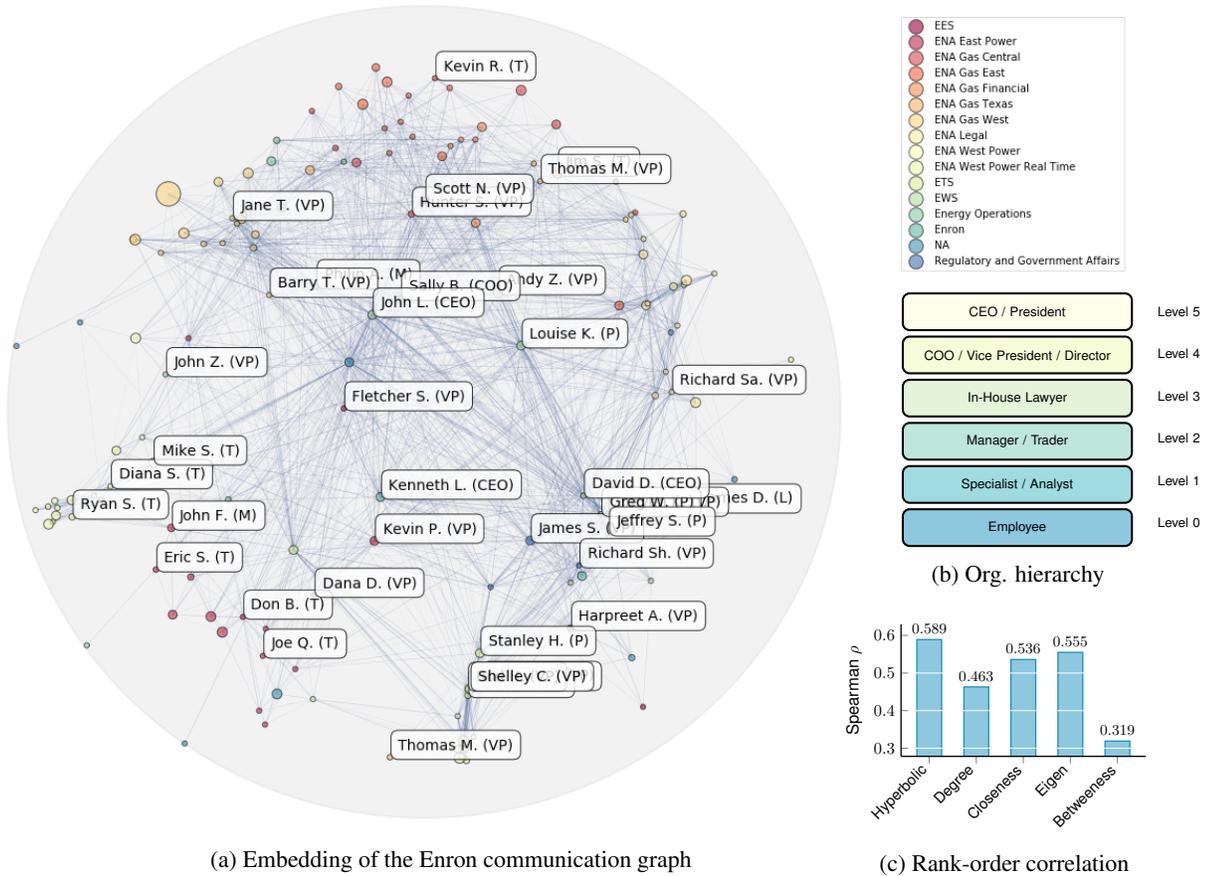
\begin{figure*}[t]
  \begin{minipage}{0.7\linewidth}
  \centering
  \includegraphics[clip,trim={2cm 1cm 0.1cm
    1cm},width=1.25\linewidth]{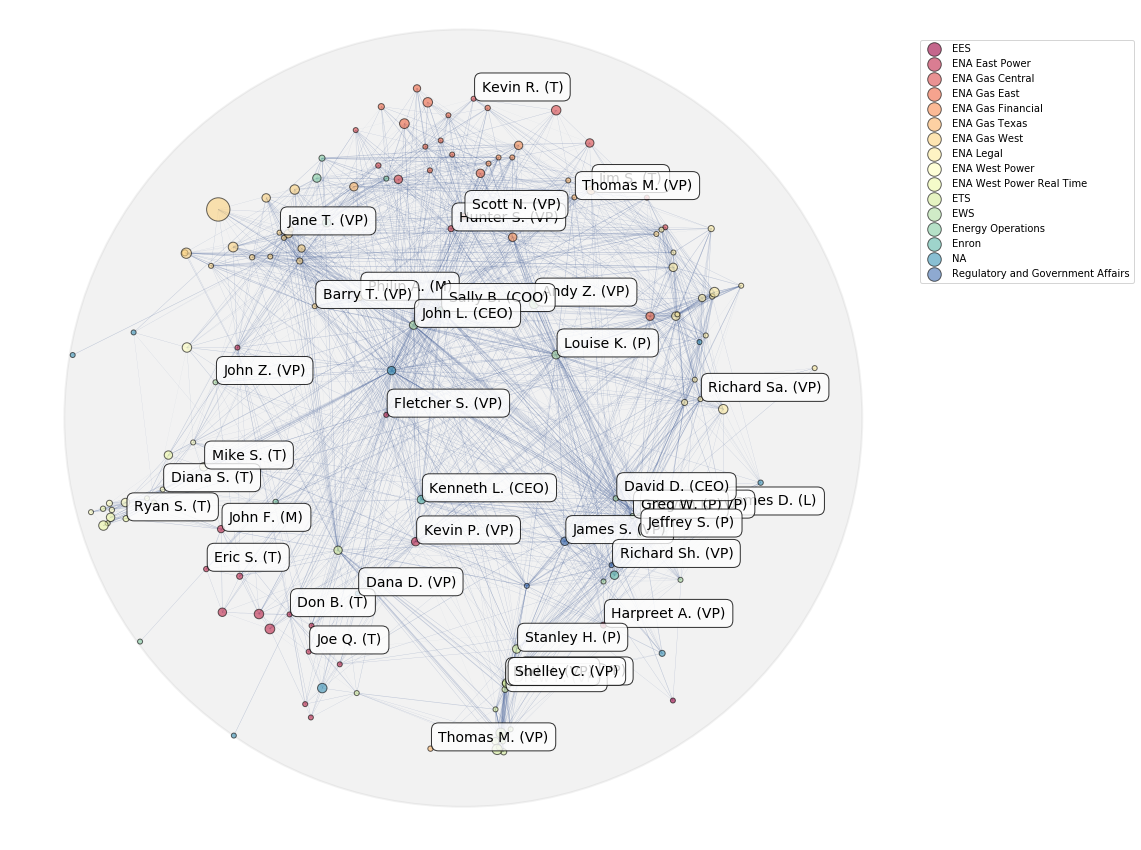}
  \subcaption{Embedding of the Enron communication graph}\label{fig:enron:embedding}
\end{minipage}%
\begin{minipage}{0.3\linewidth}
  \vspace{11em}
  
  \hspace{1.5em}
  \begin{minipage}{.7\linewidth}
  \resizebox{\linewidth}{!}{
    \hspace{-3em}\begin{tikzpicture}[
    scale=0.5,
    start chain=1 going below, 
    start chain=2 going left,
    node distance=1mm,
    desc/.style={
      scale=0.75,
      on chain=2,
      rectangle,
      rounded corners,
      draw=black, 
      very thick,
      text centered,
      text width=6cm,
      minimum height=10mm,
      fill=blue!30
		},
    it/.style={
      fill=blue!10
    },
    level/.style={
      scale=0.75,
      on chain=1,
      minimum height=10mm,
      text width=2cm,
      text centered
    },
    every node/.style={font=\sffamily}
    ]

    \node [level] (Level 5) {Level 5};
    \node [level] (Level 4) {Level 4};
    \node [level] (Level 3) {Level 3};
    \node [level] (Level 2) {Level 2};
    \node [level] (Level 1) {Level 1};
    \node [level] (Level 0) {Level 0};

    \chainin (Level 5); 
    \node [desc, fill=YlGnBu-9-1!50] (CEO) {CEO / President};
    \node [desc, fill=YlGnBu-9-2!50, continue chain=going below] (ERP) {COO / Vice President / Director};
    \node [desc, fill=YlGnBu-9-3!50] (Operations) {In-House Lawyer};
    \node [desc, fill=YlGnBu-9-4!50] (Supervisory) {Manager / Trader};
    \node [desc, fill=YlGnBu-9-5!50] (PLC) {Specialist / Analyst};
    \node [desc, fill=YlGnBu-9-6!50] (SIS) {Employee};
  \end{tikzpicture}}
  \captionsetup[sub]{oneside,margin={-5em,0cm}}
  \subcaption{Org. hierarchy}\label{fig:enron:hierarchy}
  \end{minipage}

  \vspace{1em}

  \begin{minipage}{.8\linewidth}
  \hspace{-2.5em}%
  \resizebox{\linewidth}{!}{
    \begin{tikzpicture}[thick,font=\sffamily]
      \begin{axis}[
        ybar,
        x=1cm,
        y=8cm,
        axis x line*=bottom,
        axis y line*=none,
        bar width=1.5em,
        ymajorgrids,
        major grid style=white,
        axis on top,
        thick,
        nodes near coords={
          \footnotesize\color{black}\pgfmathprintnumber[precision=3]{\pgfplotspointmeta}
        },
        enlargelimits=0.15,
        ylabel={Spearman $\rho$},
        symbolic x coords={Hyperbolic, Degree, Closeness, Eigen, Betweeness},
        xtick=data,
        x tick label style={rotate=45,anchor=east,font=\small\sffamily},
        ]
        \addplot+[ybar, thick, draw=YlGnBu-9-6,fill=YlGnBu-9-6!50] plot coordinates {(Hyperbolic,0.589) (Degree,0.463) 
          (Closeness,0.536) (Eigen,0.555) (Betweeness,0.319)};
      \end{axis}
    \end{tikzpicture}}
    \captionsetup[sub]{oneside,margin={-4em,0cm}}
    \subcaption{Rank-order correlation}\label{fig:enron:ranking}
  \end{minipage}
  \end{minipage}
  \caption{\label{fig:enron}
    Embedding of the Enron email corpus. Abbreviations in parentheses indicate organizational role: CEO = Chief Executive Officer, COO = Chief Operating Officer, P = President, VP = Vice President, D = Director, M = Manager, T = Trader. Blue lines indicate edges in the graph. Node size indicates node degree.}
\end{figure*}

In addition to the taxonomies in \Cref{sec:exp:tax}, we are interested in
discovering hierarchies from real-world graphs that have not been generated from
a clean DAG. For this purpose, we embed the communication graph of the Enron
email corpus \citep{priebe2006enron} which consists of 125,409 emails that have
been exchanged between 184 email addresses and 150 unique users.\footnote{This
dataset has been created by \citet{priebe2006enron} from the full Enron email
corpus which has been released into public domain by the Federal Energy
Regulatory Commission (FERC).} From this data, we construct a graph where
weighted edges represent the total number of emails that have been exchanged
between two users. The dataset includes also the organizational roles for 130
users, based largely on the information collected by \citet{shetty2005enron}.

\Cref{fig:enron:embedding} shows the two-dimensional embedding of this graph. It
can be seen that the embedding captures important properties of the
organizational structure. First, the nodes are approximately arranged according
to the organizational hierarchy of the company. Executive roles such as CEOs,
COOs, and (vice) presidents are embedded close to the origin, while other
employees (e.g., traders and analysts) are located closer to the boundary.
\Cref{fig:enron:ranking} shows the Spearman correlation of the norms of
the embedding with the organizational rank. It can be seen that the
norm correlates well with the ground-truth ranking and is on-par or better than
commonly-used centrality measures on graphs. We also observe that the embedding
provides a meaningful clustering of users. For instance, the lower left of the disk
shows a cluster of traders. Above that cluster (i.e.,
closer to the origin), are managers (e.g., John F.) and vice presidents (e.g.,
Fletcher S., Kevin P.) who have been associated with the trading arm of Enron.
This illustrates that, in addition to the notion of rank in a hierarchy, the
embedding provides also insight into the similarity of nodes within the
hierarchy.

\subsection{Historical Linguistics Data}
\label{sec:orgcb151c0}
\begin{figure*}[t]
  \centering
  \includegraphics[clip,trim={2cm 1cm 0cm 1cm},width=0.8\linewidth]{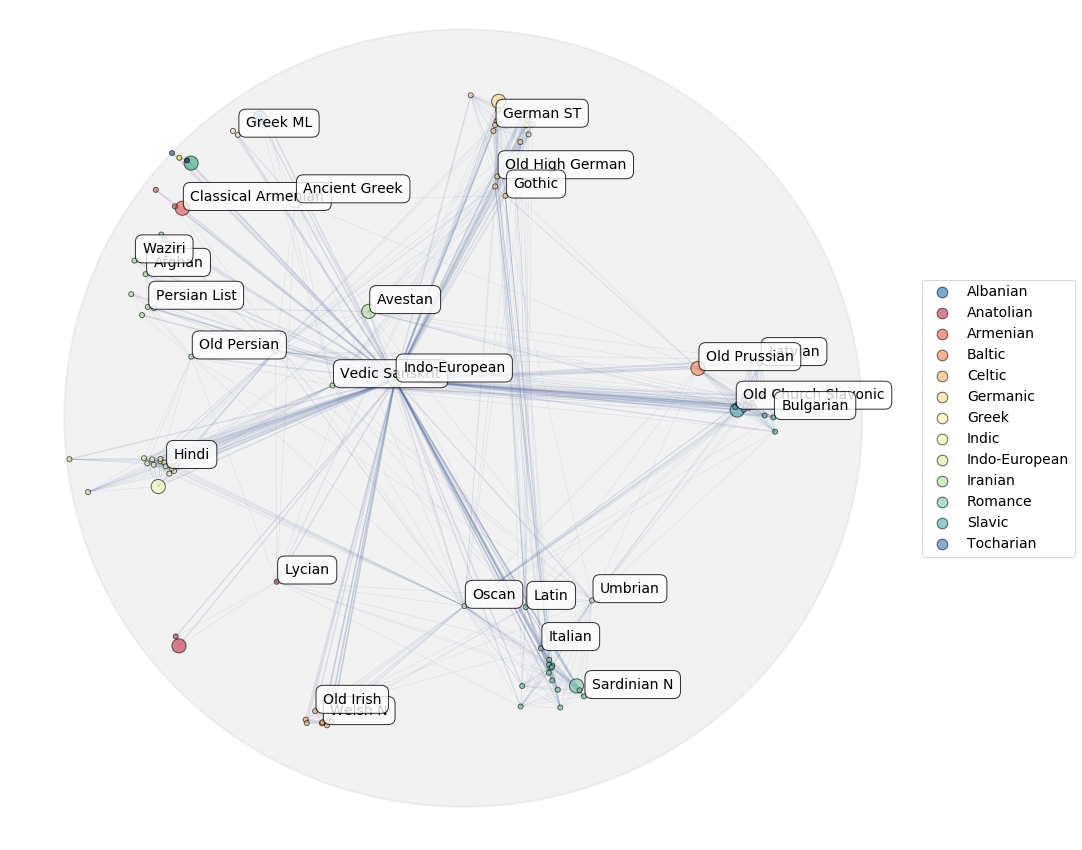}
  \caption{\label{fig:ielex}
    Embedding of the IELex lexical cognate data.}
\end{figure*}

The field of historical linguistics is concerned with the history of languages,
their relations, and their origins. An important concept to determine the
relations between languages are so-called \emph{cognates}, i.e., words that are
shared across different languages (but not borrowed) and which indicate common
ancestry in the history of languages. To be classified as cognate, words must
have similar meaning and systematic sound correspondences. Languages are assumed
to be related if they share a large number of cognates.

The goal of our experiments was to discover the historical relationships between
languages (which are assumed to follow a hierarchical tree structure) by embedding cognate
similarity data. For this purpose, we used the lexical cognate data provided by
\citet{bouckaert2012mapping}, which consists of 103 Indo-European languages and
6280 cognate sets in total. Since the number of cognate sets grew over time, not
all languages are annotated with all possible sets. For this reason, we computed
the cognate similarity between two languages in the following way. Let
\(c(\ell_1, \ell_2)\) denote the number of common cognates in languages \(\ell_1,
\ell_2\). Furthermore, let \(a(\ell_1)\) denote the number of cognate annotations
for \(\ell_1\). We then compute the cognate similarity of \(\ell_1, \ell_2\) simply
as
\begin{equation*}
  \text{csim}(\ell_1, \ell_2) = \frac{c(\ell_1, \ell_2)}{\min(a(\ell_1), a(\ell_2))} .
\end{equation*}

\Cref{fig:ielex} shows a two-dimensional embedding of these cognate similarity
scores. It can be seen that the embedding allows us to discover a meaningful
hierarchy that corresponds well with the assumed origin of languages. First, the
embedding shows clear clusters of high-level language families such as Celtic, Romance,
Germanic, Balto-Slavic, Hellenic, Indic, and Iranian. Moreover, each of these
cluster displays meaningful internal hierarchies such as (Gothic \(\to\) Old High
German \(\to\) German), (Old Prussian \(\to\) Old Church Slavonic \(\to\) Bulgarian),
(Latin \(\to\) Italian), or (Ancient Greek \(\to\) Greek). Closer to the center of the
disc, we also find a number of ancient languages. For instance, Oscan and
Umbrian are two extinct sister languages of Latin and located above the Romance
cluster, Similarly, Avestan and Vedic-Sanskrit are two ancient languages that
separated early in the pre-historic era before 1800 BCE
\citep{baldi1983introduction}. After separation, Avestan developed in ancient
Persia while Vedic-Sanskrit developed independently in ancient India. In the
embedding, both languages are close to the center and to each other.
Furthermore, Avestan is close to the Iranian cluster while Vedic-Sanskrit is
close to the Indic cluster.

\section{Conclusion}
\label{sec:org67c037e}
We introduced a new method for learning continuous concept hierarchies from
unstructured observations. We exploited the properties of hyperbolic geometry in
such a way that we can discover hierarchies from pairwise similarity scores --
under the assumption that concepts in the same subtree of the ground-truth
hierarchy are more similar to each other than to concepts in different subtrees.
To learn the embeddings, we developed an efficient Riemannian optimization
approach based on the Lorentz model of hyperbolic space. Due to the more
principled optimization approach, we were able to substantially improve the
quality of the embeddings compared to the method proposed by
\citet{nickel2017poincare} -- especially in low dimensions. We further showed on
two real-world datasets, that our method can discover meaningful hierarchies
from nothing but pairwise similarity information.

\subsection*{Acknowledgments}
\label{sec:orge0370d4}
The authors thank Joan Bruna, Martín Arjovsky, Eryk Kopczyński, and Laurens van der
Maaten for helpful discussions and suggestions.

\bibliographystyle{icml2018}
\bibliography{icml18_references}
\end{document}

%% file: geodesics.pgf
\pgfdeclarelayer{background}
\pgfdeclarelayer{foreground}
\pgfsetlayers{background,main,foreground}
\begin{tikzpicture}[very thick,scale=0.5]
\begin{pgfonlayer}{foreground}\draw (0,0) circle (3.0);\end{pgfonlayer}
\tikzstyle{segment}=[line width=0.3mm]

\begin{pgfonlayer}{background}
\end{pgfonlayer}

\draw[black] (-1.142, 2.774) arc (22.383:-22.383:7.285);
\draw[black] (-0.333, 2.981) arc (-173.619:-36.381:1.175);
\draw[black] (-1.500, -2.598) -- (1.500, 2.598);
\draw[segment,magenta] (0.000, 0.000) -- (1.142, 1.979);
\draw[segment,orange] (0.000, 2.285) arc (-135.230:-74.770:1.175);
\draw[segment,cyan] (-0.693, 1.201) arc (9.486:-9.486:7.285);

\begin{pgfonlayer}{foreground}
\draw[fill=black,black] (0.000,0.000) circle (0.05);
\draw[fill=black,black] (-0.693,-1.201) circle (0.05);
\draw[fill=black,black] (1.142,1.979) circle (0.05);
\draw[fill=black,black] (-0.693,1.201) circle (0.05);
\draw[fill=black,black] (0.000,2.285) circle (0.05);
\node at (0.4,-0.2) {\tiny $p_1$};
\node at (-0.3,-1.4) {\tiny $p_2$};
\node at (1.55,1.77) {\tiny $p_3$};
\node at (-1,1) {\tiny $p_4$};
\node at (-0.2,2) {\tiny $p_5$};
\end{pgfonlayer}
\end{tikzpicture}